\documentclass[letterpaper, 10 pt, conference]{ieeeconf}  
\usepackage{csquotes}

\IEEEoverridecommandlockouts                              

\usepackage[T1]{fontenc}
\usepackage{graphics} 
\usepackage{graphicx}
\usepackage{epsfig} 
\usepackage{amsmath} 
\usepackage{amssymb}  
\usepackage{booktabs}
\usepackage{tikz}
\usepackage{multirow}
\usepackage[table]{xcolor}
\usepackage{cite}

\makeatletter
\let\NAT@parse\undefined
\makeatother

\usepackage[colorlinks]{hyperref}
\hypersetup{
  citecolor=blue,
  linkcolor=blue,
  urlcolor=blue
}
\title{\LARGE \bf
Stein Variational Ergodic Surface Coverage with SE(3) Constraints
} 

\author{Jiayun Li$^{1,2}$, Yufeng Jin$^{1,3}$, Sangli Teng$^{4}$, Dejian Gong$^{1}$, Georgia Chalvatzaki$^{1,2}$
\thanks{Corresponding author: Jiayun Li (jiayun.li@tu-darmstadt.de).}
\thanks{Yufeng Jin and Sangli Teng contributed equally to this work.}%
\thanks{This work was supported by the German Research Foundation (DFG) Emmy Noether Programme (CH 2676/1-1), the EU’s Horizon Europe project \enquote{ARISE} (Grant no.: 101135959), the German Federal Ministry of Education and Research (BMBF) Project \enquote{RiG} (Grant no.: 16ME1001).}
\thanks{$^{1}$Department of Computer Science, TU Darmstadt, Darmstadt, Germany.}%
\thanks{$^{2}${Hessian.AI}, Darmstadt, Germany.}%
\thanks{$^{3}$Honda Research Institute Europe GmbH, Offenbach, Germany.}%
\thanks{$^{4}$University of California, Berkeley, USA.}%
}

\begin{document}

\maketitle
\thispagestyle{empty}
\pagestyle{empty}

\begin{abstract}
Surface manipulation tasks require robots to generate trajectories that comprehensively cover complex 3D surfaces while maintaining precise end-effector poses. Existing ergodic trajectory optimization (TO) methods demonstrate success in coverage tasks, while struggling with point-cloud targets due to the nonconvex optimization landscapes and the inadequate handling of SE(3) constraints in sampling-as-optimization (SAO) techniques. In this work, we introduce a preconditioned SE(3) Stein Variational Gradient Descent (SVGD) approach for SAO ergodic trajectory generation. Our proposed approach comprises multiple innovations. First, we reformulate point-cloud ergodic coverage as a manifold-aware sampling problem. Second, we derive SE(3)-specific SVGD particle updates, and, third, we develop a preconditioner to accelerate TO convergence. Our sampling-based framework consistently identifies superior local optima compared to strong optimization-based and SAO baselines while preserving the SE(3) geometric structure. Experiments on a 3D point-cloud surface coverage benchmark and robotic surface drawing tasks demonstrate that our method achieves superior coverage quality with tractable computation in our setting relative to existing TO and SAO approaches, and is validated in real-world robot experiments. 
\smallskip
\noindent\textbf{Code, data, and video:} 
\href{https://github.com/JonasPflaume/tsvec}{https://github.com/JonasPflaume/tsvec}
\end{abstract}

\section{INTRODUCTION}

Trajectory optimization (TO) for robots operating on complex surfaces with vision feedback represents a fundamental challenge in modern robotics, with critical applications spanning manufacturing automation \cite{maric2020collaborative}, surgical robotics \cite{fu2024optimization}, and autonomous maintenance systems \cite{pfandler2024non}. In such scenarios, robots must generate motion plans that simultaneously satisfy kinematic constraints, achieve comprehensive coverage of task-relevant regions, and maintain precise geometric relationships between the end-effector and surface geometry.

Ergodic TO provides a principled mathematical framework for addressing this challenge by generating trajectories that asymptotically visit regions in proportion to their information content \cite{mathew2011metrics}. Unlike myopic exploration strategies that focus on immediate information gain \cite{pathak2017curiosity}, ergodic methods optimize the long-term spatial distribution of robot visitation, ensuring both coverage completeness and task-oriented exploration.

However, when applied to discrete point-cloud targets with SE(3) end-effector constraints, existing ergodic TO methods face three critical limitations. First, point-cloud ergodic coverage formulations generate highly non-convex optimization landscapes with numerous local minima that trap continuous optimization-based planners. Second, while ergodic TO has been extended to SE(3) spaces \cite{sun2024fast}, existing Sampling As Optimization (SAO) methods using sampling or flow matching \cite{sun2025flow} designed to overcome local optima fail to properly handle the SE(3) manifold geometry, limiting their applicability to surface manipulation tasks which require very precise pose control. Third, flow-based and score-based SAO methods suffer from severe ill-conditioning in long-horizon TO problems, while effective preconditioning strategies for these approaches remain largely unexplored.

To address these limitations, we develop a theoretically grounded preconditioned SE(3) Stein Variational Gradient Descent (SVGD) framework for SAO ergodic TO with discrete point-cloud input. Our approach integrates ergodic optimization metrics with manifold-aware SVGD to enable parallel, geometrically consistent trajectory generation for surface manipulation tasks. We refer to this method as \textbf{T}ask-space \textbf{S}tein \textbf{V}ariational \textbf{E}rgodic \textbf{C}overage (TSVEC).

\begin{figure}[t]
    \centering
    \includegraphics[width=0.99\linewidth,height=0.4\linewidth]{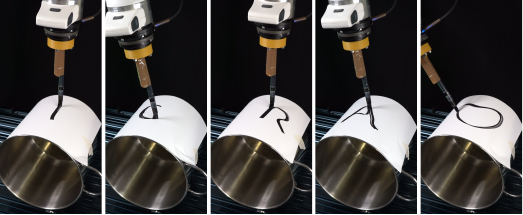}
    \caption{Robot drawing the letters 'ICRA' followed by a heart symbol on the cylindrical surface of a pot using a pen mounted on its end effector, guided by point cloud input.}
    \label{fig:real_world_panda}
\end{figure}

Our key contributions are as follows.
\begin{itemize}
\item \textbf{SVGD on the SE(3) manifold:} We derive a principled extension of SVGD to the SE(3) manifold.

\item \textbf{Preconditioned ergodic TO:} We reformulate surface ergodic coverage as a nonlinear least-squares problem and introduce a Gauss-Newton (GN) preconditioner compatible within the SE(3) SVGD framework.

\item \textbf{Comprehensive experimental validation:} We benchmark the proposed framework against strong baselines, showing consistent performance gains, and further validate its effectiveness in real-world robotic experiments.
\end{itemize}

\section{RELATED WORK}

\subsection{Ergodic Coverage and Trajectory Optimization}

Ergodic TO was introduced by Mathew and Mezić \cite{mathew2011metrics} as a framework for generating trajectories that visit regions proportional to a target information distribution. Miller et al. \cite{miller2015ergodic} demonstrated ergodic exploration for distributed information gathering, while subsequent work extended the approach to SE(2) spaces \cite{miller2013se2} and developed TO formulations \cite{miller2013trajectory}.

A key challenge in ergodic TO is the initialization sensitivity problem. Current point-cloud formulations create highly nonconvex optimization landscapes where distant regions of interest cannot effectively attract trajectories from poor initializations \cite{ayvali2017ergodic}. This stems from the spectral decomposition of the ergodic metric, which creates numerous local minima that trap continuous optimization-based optimizers. Heat equation-driven approaches, like HEDAC \cite{ivic2016ergodicity}, attempt to address this through diffusion-based information propagation, but perform poorly in continuous TO scenarios due to objective function imbalances that leave the optimization heavily influenced by initialization, failing to overcome the attraction basins of local minima \cite{hughes2025ergodic}.

Recent advances have extended ergodic methods to higher-dimensional spaces and SAO-based approaches. Sun et al. \cite{sun2024fast} developed fast ergodic search with kernel functions and demonstrated applications on SE(3) spaces. However, their approach focuses on single-point estimation with continuous target distributions and lacks integration with parallel sampling frameworks, necessary for exploring multiple trajectory modes. Sampling methods using SVGD \cite{lee2024stein} are computationally slow and lack awareness of SE(3) manifold constraints. Flow-based ergodic criteria using Sinkhorn divergence or Maximum Mean Discrepancy can handle multiple modes \cite{sun2025flow,hughes2025ergodic}, but are prohibitively expensive for TO with large point-cloud inputs, being limited to sparse subsampled representations. Safety-critical extensions \cite{lerch2022safety} and tactile surface exploration \cite{bilaloglu2025tactile} have demonstrated the versatility of ergodic methods, but computational scalability remains a fundamental limitation, restricting applications to control synthesis rather than full trajectory optimization. 

\subsection{Trajectory Optimization as Inference on SE(3)}
The formulation of TO as probabilistic inference represents a fundamental paradigm shift in motion planning. Attias \cite{attias2003planning} pioneered this approach, followed by Toussaint \cite{toussaint2009robot} who established the foundation for continuous robot trajectory optimization. CHOMP (Covariant Hamiltonian Optimization for Motion Planning) \cite{zucker2013chomp} and GPMP (Gaussian Process Motion Planning) \cite{mukadam2018continuous} further advanced this framework using functional gradients and factor graphs, respectively. While effective in Euclidean spaces, their extension to SE(3) manifolds remains challenging due to the complex geometric constraints.

SE(3) TO faces fundamental limitations in existing approaches. GPMP extensions to Lie groups \cite{dong2018sparse} operate primarily in logarithm tangent space using locally linear approximations, which ignore the intrinsic manifold structure of SE(3) and can lead to accumulation of linearization errors over long horizons \cite{blanco2021tutorial,wagner2011rapid}. This tangent space treatment fails to properly handle the non-commutative nature of rotations and parallel transport operations essential for SE(3) geometry. While recent work on Riemannian direct optimization \cite{teng2025riemannian} for Lie group trajectories shows efficiency gains, its integration with SVGD frameworks remains unclear.

\subsection{Equality-constrained / Manifold SVGD}
Stein Variational Gradient Descent (SVGD) \cite{liu2016stein} provides a non-parametric approach for approximating complex distributions through particle-based variational inference. Although there exist manifold extensions, including Riemannian SVGD \cite{liu2018riemannian} for general Riemannian manifolds, projected SVGD \cite{chen2020projected} for subspace constraints, and orthogonal-gradient SVGD \cite{zhang2022sampling} for equality constraints used in TO applications \cite{power2024constrained}, none of these are specifically designed for the SE(3) Lie group structure. These general manifold methods suffer from computational inefficiency when applied to SE(3) TO, as they fail to exploit the specific geometric properties and computational shortcuts available for matrix Lie groups.

Our proposed TSVEC addresses these limitations through a direct SE(3) manifold formulation of SVGD that exploits the manifold structure, combined with a trajectory-specific metric preconditioner, to enable robust exploration while maintaining computational efficiency on large point clouds.
\section{PRELIMINARY}
\subsection{Ergodic Coverage Optimization Problem}
\label{subsec:ergodic_prelim}
\begin{figure}[t]
    \centering
    \vspace{2mm}
    \includegraphics[width=\linewidth]{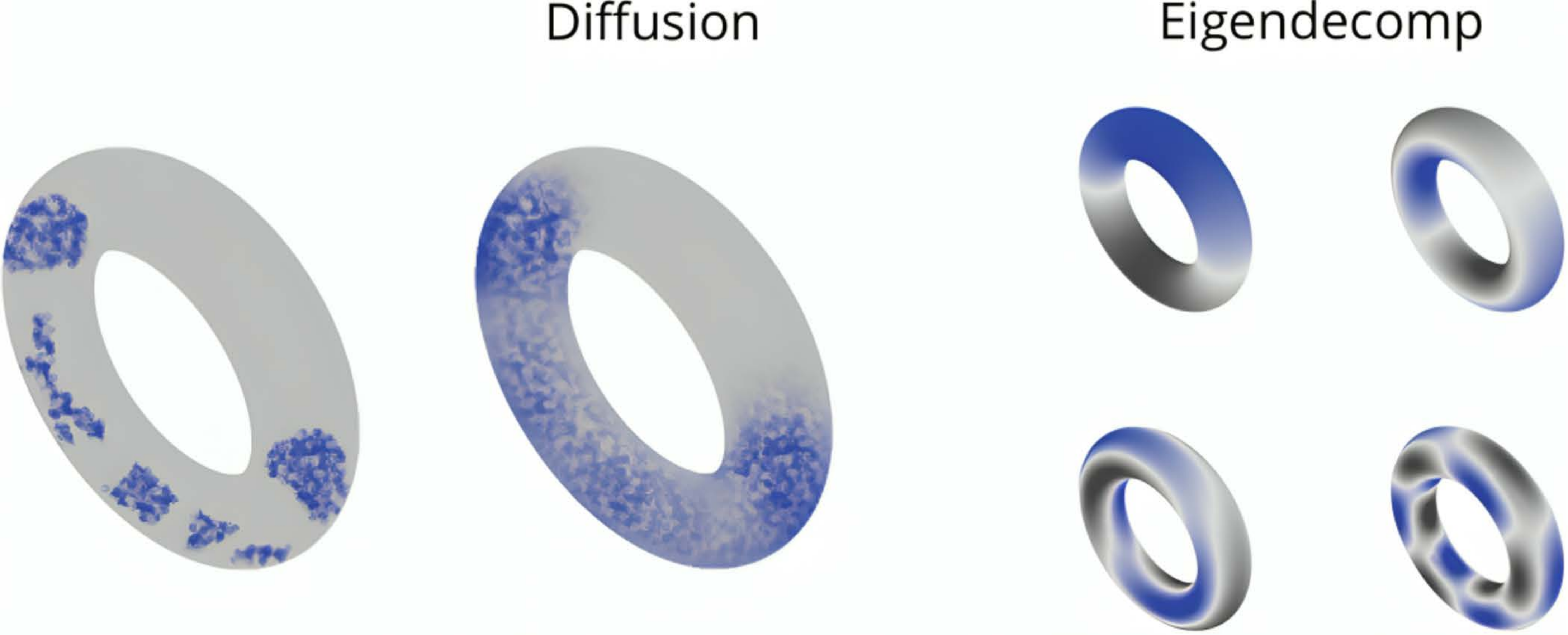}
    \caption{From left to right, the figure shows a colored point cloud on a torus model, the results after graph diffusion, and the projection onto a graph Fourier basis.}
    \vspace{-3mm}
    \label{fig:spectral}
\end{figure}

A central component of ergodic TO is the \emph{ergodic metric}, which measures the discrepancy between the spatial statistics of a trajectory and a target trajectory distribution. The target trajectory distribution $\varphi(\mathbf{x})$ is first encoded in the leading $S$ Fourier basis as coefficients
\begin{equation}
    \mathbf{c}_i = \int_{\Pi} \varphi(\mathbf{x}) F_i(\mathbf{x}) \, d\mathbf{x}, \quad i=1,\dots,S ,
\end{equation}
where $F_i(\mathbf{x})$ are cosine Fourier basis functions defined over the bounded workspace $\Pi \subset \mathbb{R}^d$.  
Similarly, the time-averaged trajectory statistics are represented in the same basis as $c(\mathbf{X})_i$.
With this representation, the ergodic objective can be written as a nonlinear least-squares problem:
\begin{equation}
    V_e(\mathbf{X}) = \frac{1}{2} \sum_{i=1}^{S} \Lambda_i \big( c(\mathbf{X})_i - \mathbf{c}_i \big)^2 ,
    \label{eq:ergodic_metric_prelim}
\end{equation}
where $\Lambda_i$ are frequency-dependent weights. Minimizing $V_e(\mathbf{X})$ enforces that the trajectory $\mathbf{X} = \{\mathbf{x}_t \in \text{SE}(3)\}_{t=1}^{N_t}$, of length $N_t$, becomes ergodic with respect to the target distribution, i.e., that the time spent in each region is proportional to its utility. This formulation retains a nonlinear and nonconvex structure but is amenable to TO methods \cite{miller2015ergodic, lerch2022safety}. It is worth noting that for discrete domains such as point clouds, one can first diffuse the point cloud to build a smooth distribution, then adopt the graph Fourier transform, i.e., the eigendecomposition of the discrete Laplace–Beltrami operator; Fig. \ref{fig:spectral} shows a diffusion example on a torus; see \cite{bilaloglu2025tactile} for further details. 

\subsection{Stein Variational Inference}
SVGD updates a set of $N$ particles $\{x_i\}$ so that their distribution progressively matches a target distribution $p(x)$, possibly unnormalized. Rather than directly optimizing the KL divergence in parameter space, SVGD finds the direction that decreases KL the fastest within a function space defined by a kernel $k(x,y)$, resulting in a kernelized update that transports particles toward the target. This yields a closed-form update rule $\phi^*$ which is a vector field shown in Eq.~(\ref{eq:svgd}) that moves each particle $x_i$ according to both the kernel-smoothed target distribution's score $\nabla \log p(x_i)$ and a repulsive kernel force $\nabla_1 k(x_i, x)$. We use the notation $\nabla_{i}$ to denote the gradient with respect to the $i$-th variable for a multi-input function. The repulsive force can be understood as the maximum entropy term in the general variational inference framework. 
\begin{align}
\label{eq:svgd}
\phi^*(x) &= \frac{1}{N} \sum_{i=1}^N \Big[ k(x_i, x) \, \nabla \log p(x_i) + \nabla_{1} k(x_i, x) \Big]
\end{align}

\subsection{SE(3) Preliminary}
To derive SVGD on SE(3), we need the algebraic and differential operations on this manifold. We consider perturbations of the special Euclidean group $\mathrm{SE}(3)$ in the form of right multiplication. For a matrix $\mathbf{P} \in \mathrm{SE}(3)$, we define the perturbation as $\mathbf{P} \oplus \boldsymbol{\epsilon} = \mathbf{P} \exp(\boldsymbol{\epsilon}^{\wedge})$, where $\boldsymbol{\epsilon} \in \mathbb{R}^{6}$ is a small perturbation vector, $(\cdot)^{\wedge}$ denotes the hat operator mapping from $\mathbb{R}^{6}$ to the Lie algebra $\mathfrak{se}(3)$, and $\exp$ is the matrix exponential. Similarly, we can define SE(3) subtraction as $\mathbf{P}' \ominus \mathbf{P} = \log(\mathbf{P}^{-1}\mathbf{P}')^{\vee} \in \mathbb{R}^6$, where the vee operator $(\cdot)^{\vee}$ maps from $\mathfrak{se}(3)$ to $\mathbb{R}^{6}$. We use Ad and ad to denote the (group/algebra) adjoint operators on SE(3).

For a smooth function $F$ that maps SE(3) to SE(3), its right-perturbed Jacobian can be written as a matrix $\mathbf{J} F(\mathbf{P}) \in \mathbb{R}^{6 \times 6}$:
\begin{equation}
\label{eq:transport_jacobian}
\mathbf{J} F(\mathbf{P}) = \lim_{\boldsymbol{\epsilon} \to \mathbf{0}} \frac{ F(\mathbf{P} \oplus \boldsymbol{\epsilon}) \ominus F(\mathbf{P}) }{\boldsymbol{\epsilon}},
\end{equation}
where the limit is taken component-wise, we refer to \cite{sola2018micro} for more details on SE(3), including the chain rule and basic Lie algebra operations. We consider the left-invariant inner product on the tangent space $\mathcal{T}_{\mathbf{P}}\mathrm{SE}(3) \cong \mathfrak{se}(3)$ as $ \langle \mathbf{P}\xi_1, \mathbf{P}\xi_2  \rangle_{\mathbf{P}} = \langle \xi_1, \xi_2 \rangle_{\mathbf{P}} = \frac{1}{2}\mathrm{Tr}(\xi_1^{\wedge} (\xi_2^{\wedge})^{\top})$ for $\xi_1^{\wedge}, \xi_2^{\wedge} \in \mathcal{T}_{\mathbf{P}}\mathrm{SE}(3)$. This right-perturbation model naturally preserves the group structure and provides a consistent framework for optimization on $\mathrm{SE}(3)$. Since SE(3) is non-compact, there is, in general, no bi-invariant Riemannian metric, so the Riemannian exponential need not coincide with the Lie exponential. Nevertheless, one can perform second-order optimization by working in canonical coordinates and using the Cartan–Schouten connections to define gradients/Hessians; this yields Newton methods with local quadratic convergence using only the expansion of Lie exponentials, see \cite{mahony2002geometry}. 

\section{METHOD}

\subsection{Stein Variational Gradient Descent on SE(3)}
To derive SVGD on $SE(3)$ from first principles, we define a small perturbation (i.e., a transport map $T$) applied to particles on $SE(3)$ that represent the variational distribution $q$. One natural choice is to directly adopt the SE(3) right addition $T(\mathbf{P}) = \mathbf{P} \oplus \tau \phi(\mathbf{P})$. Here, $\phi: SE(3) \rightarrow \mathbb{R}^6$ denotes the SVGD vector field on $SE(3)$, and the updates are performed with a small step size $\tau$. We follow the standard approach to derive SVGD by examining the steepest descent direction of the perturbed KL divergence:
\begin{equation}
\label{eq:perturbed_KL}
    \mathrm{KL}\!\left(q_{[T]} \,\|\, p \right) = \mathbb{E}_{\mathbf{P} \sim q} \!\left[ \log q(T(\mathbf{P})) - \log p(\mathbf{P}) + \log Z\right],
\end{equation}
where $q_{[T]}$ denotes the variational distribution after applying the transport map $T$, and $Z$ is the normalization constant, which is invariant under perturbations and can therefore be neglected. Starting from equation (\ref{eq:perturbed_KL}), we apply the change of variables formula to transform the analysis from the particle variational distribution to the unnormalized posterior distribution. Specifically, by applying $T^{-1}$ to the particles, we can work with the inverse perturbation that directly acts on the target distribution rather than the variational distribution. This transformation allows us to derive the steepest descent direction with respect to the target distribution, which can be shown to be:
\begin{align}
       \nabla_{\tau} \text{KL}(q_{[T]}||p) &= - \mathbb{E}_{\mathbf{P} \sim q} \!\left[ \nabla_{\tau} \log p_{[T^{-1}]}(\mathbf{P}) \!\right],
\end{align}
where the terms inside the expectation are
\begin{equation}
\label{eq:grad_inverse}
    \nabla_{\tau} \log p_{[T^{-1}]}(\mathbf{P}) = \nabla_{\tau} \log p(T(\mathbf{P})) + \nabla_{\tau} \log |\text{det}(\mathbf{J} T(\mathbf{P}))|.
\end{equation}
These derivations follow closely the supplementary material of \cite{liu2016stein}. 

The first term in Eq. (\ref{eq:grad_inverse}) is equal to the gradient of the log-likelihood function $\text{grad} \log p(\mathbf{P})$. 
For an objective function $f(\mathbf{P})$ with domain on $\mathrm{SE}(3)$, the gradient with respect to right perturbations is computed by first obtaining the Euclidean gradient $\nabla f(\mathbf{P}) \in \mathbb{R}^{4 \times 4}$ and then projecting it onto the tangent space $\mathcal{T}_{\mathbf{P}}\mathrm{SE}(3)$. Given the definition of the inner product, the following directional derivative holds:
\begin{equation}
    \nabla_\tau f(\mathbf{P}\oplus \tau \boldsymbol{\epsilon})\bigg|_{\tau=0} = \langle \mathrm{grad} f(\mathbf{P}), \mathbf{P}\boldsymbol{\epsilon}^{\wedge}\rangle_{\mathbf{P}},
\end{equation}
where $\boldsymbol{\epsilon} \in \mathbb{R}^6$ is a perturbation vector. The gradient $\mathrm{grad} f(\mathbf{P})$ can be computed by:
\begin{equation}
\label{eq:riem_gradient}
\mathrm{grad} f(\mathbf{P}) = \mathbf{P} 
\begin{bmatrix}
\boldsymbol{\Omega} & \mathbf{R}^{\top} \dfrac{\partial f}{\partial \mathbf{r}} \\[0.5em]
\mathbf{0}_{1 \times 3} & 0
\end{bmatrix},
\end{equation}
where 
\begin{equation}
\boldsymbol{\Omega} = \mathbf{R}^{\top} \frac{\partial f}{\partial \mathbf{R}} - \frac{\partial f}{\partial \mathbf{R}}^{\top} \mathbf{R}
\end{equation}
and $\mathbf{P}$ is parameterized as:
\begin{equation}
\mathbf{P} = \begin{bmatrix} 
\mathbf{R} & \mathbf{r} \\[0.3em] 
\mathbf{0}_{1 \times 3} & 1 
\end{bmatrix}
\end{equation}
with $\mathbf{R} \in \mathrm{SO}(3)$ the orientation and $\mathbf{r} \in \mathbb{R}^3$ the position. For further technical details about the gradient of SE(3) functions, see \cite{barfoot2024state} (Section 8.1) and \cite{boumal2023introduction}.

The second term involving the derivative of the log-determinant can be expressed using the standard identity:
\begin{align}
    \nabla_{\tau} \log |\text{det}(\mathbf{J} T(\mathbf{P}))| = \rm{Tr} [\mathbf{J} T(\mathbf{P})^{-1} \nabla_{\tau} \mathbf{J} T(\mathbf{P})]
\end{align}
The $\mathbf{J} T(\mathbf{P})$ approaches the identity for sufficiently small $\tau$; however, its derivative with respect to $\tau$ is not constant. One can show this by using the definition of the transport map Jacobian in Eq. (\ref{eq:transport_jacobian}) with an additional perturbation $\boldsymbol{\epsilon}$ for Jacobian computation. Firstly, we take the derivative with respect to $\tau$:
\begin{align}
     \nabla_{\tau} \mathbf{J} T(\mathbf{P})& \approx \phi(\mathbf{P} + \boldsymbol{\epsilon}^{\wedge}\mathbf{P})^{\wedge} (I + \boldsymbol{\epsilon}^{\wedge}) - (I + \boldsymbol{\epsilon}^{\wedge}) \phi(\mathbf{P})^{\wedge} \\
    &\approx [\mathbf{J} \phi(\mathbf{P}) \boldsymbol{\epsilon}]^{\wedge} + \text{Lie}[\phi(\mathbf{P})^{\wedge}, \boldsymbol{\epsilon}^{\wedge}],
\end{align}
where $\text{Lie}[\cdot]$ is the Lie bracket. Then, we extract the Jacobian by extracting the first-order term of  $\boldsymbol{\epsilon}$:
\begin{equation}
    \nabla_{\tau} \mathbf{J} T(\mathbf{P}) = \mathbf{J}\phi(\mathbf{P}) + \text{ad}_{\phi(\mathbf{P})^{\wedge}}.
\end{equation}
The trace of the $\mathfrak{se}(3)$ adjoint is 0; therefore $\nabla_{\tau} \log |\text{det}(\mathbf{J} T(\mathbf{P}))| = \text{Tr}[\mathbf{J}\phi(\mathbf{P})]$, which is the maximum entropy term of SVGD on SE(3). By using the notation of  gradient to represent the trace of the Jacobian element-wise, the steepest descent direction is
\begin{equation}
\nabla_{\tau} \text{KL}(q_{[T]}||p) = - \mathbb{E}_{\mathbf{P} \sim q}\left[\sum_{i=1}^6\langle \boldsymbol{\Phi}_i(\mathbf{P}), \mathbf{P} I_i^{\wedge}\rangle_{\mathbf{P}} \right],
\end{equation}
where $\boldsymbol{\Phi}_i(\mathbf{P}) = \phi_i(\mathbf{P}) \, \text{grad}\, \log p(\mathbf{P}) + \text{grad}\, \phi_i(\mathbf{P})$ represents the gradient contribution from the $i$-th component of the vector field $\phi$, with $\{I_i^{\wedge}\}_{i=1}^6$ forming the standard basis for $\mathfrak{se}(3)$. To obtain a closed-form solution for discrete particles, we apply the kernel trick (reproducing property) introduced in SVGD \cite{liu2016stein} to represent the vector field $\phi$ using a kernel function $k$, and we constrain the gradient norm to be unity with respect to the kernel-induced norm. We obtain the contribution of the gradient from $\mathbf{P}_i$ to $\mathbf{P}'$, as given in Eq.~(\ref{eq:psi}), when evaluating the vector field at $\mathbf{P}'$. It should be noted that the gradient is still represented at point $\mathbf{P}$, therefore, a parallel transport scheme is needed.


In the context of SVGD on SE(3) manifolds, parallel transport of tangent vectors between different tangent spaces is essential for maintaining geometric consistency when comparing or aggregating gradient information across particles. For the group SE(3), we must transport tangent vectors from one tangent space $\mathcal{T}_{\mathbf{P}}\mathrm{SE}(3)$ to another $\mathcal{T}_{\mathbf{P}'}\mathrm{SE}(3)$ to ensure that the kernel-induced interactions between particles respect the underlying manifold structure.

The parallel transport operation can be efficiently computed using the adjoint representation of the Lie group. Specifically, for a tangent vector $\boldsymbol{\xi} \in \mathcal{T}_{\mathbf{P}}\mathrm{SE}(3) \cong \mathfrak{se}(3)$, its parallel transport to $\mathcal{T}_{\mathbf{P}'}\mathrm{SE}(3)$ is given by:
\begin{equation}
    \label{eq:parallel_transport}
    \mathcal{P}_{\mathbf{P} \rightarrow \mathbf{P}'}(\boldsymbol{\xi}) = \mathrm{Ad}_{\mathbf{P}'^{-1}\mathbf{P}} \boldsymbol{\xi}
\end{equation}
where $\mathrm{Ad}_{(\cdot)}$ denotes the adjoint operator to change the basis for vectors in different tangent spaces. This formulation ensures that the transported vectors maintain their geometric meaning while accounting for the relative transformation between the two poses. Therefore, the target vector field is:
\begin{align}
\phi^*(\mathbf{P}') &= \frac{1}{N} \sum_{i=1}^{N}
   \mathcal{P}_{\mathbf{P}_i \rightarrow \mathbf{P}'}\bigl( \psi(\mathbf{P}_i, \mathbf{P}') \bigr), \label{eq:phi}\\[0.5em]
\psi(\mathbf{P}_i, \mathbf{P}') &= k(\mathbf{P}_i, \mathbf{P}')\, \mathrm{grad}\, \log p(\mathbf{P}_i)
   + \mathrm{grad}_1\, k(\mathbf{P}_i, \mathbf{P}'). \label{eq:psi}
\end{align}
\begin{figure}[t]
  \centering
  \includegraphics[width=0.6\columnwidth]{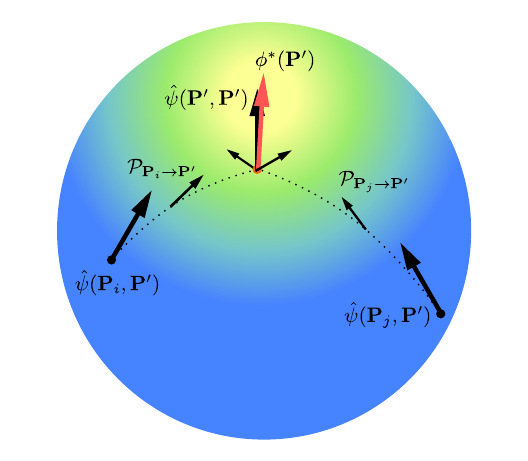}
  \caption{The SE(3) SVGD mechanism for sharing gradient information on the manifold: computing the natural gradient (black arrow) and applying parallel transport to obtain the final update direction (red arrow).}
  \vspace{-3mm}
  \label{fig:se3_svgd}
\end{figure}
The idea of SE(3) SVGD is illustrated in Fig.~\ref{fig:se3_svgd}, highlighting the role of parallel transport in aggregating gradients consistently on the manifold.

\subsection{Ergodic TO as Stein Variational Inference on SE(3)}
\label{subsec:se_3_to}
In TO tasks with long horizons, particularly in the ergodic TO setting, trajectories may span several hundred steps. This leads to severe ill-conditioning of the optimization problem, thereby necessitating the development of additional preconditioners to accelerate convergence \cite{zucker2013chomp, mukadam2018continuous}.

To perform surface ergodic trajectory optimization with SE(3) SVGD, we formulate our inference problem using a Boltzmann likelihood \(p(\mathbf{X}) \propto \exp(-V(\mathbf{X}))\), 
where the energy \(V(\mathbf{X})\) is the sum of four terms: \(V_s\) (smoothness), \(V_a\) (surface normal alignment), \(V_f\) (surface attachment), and \(V_e\) (ergodic metric) along the trajectory \(\mathbf{X} = \{\mathbf{x}_t \in \text{SE}(3)\}_{t=1}^{N_t}\). We assume that all Lie operations (e.g.,  gradient, parallel transport, wedge, and vee operators) are computed element-wise for each time step. We also assume access to a continuously differentiable signed distance function (SDF) $f_{sdf} \in C^2$ whose gradient is normalized, i.e., $\|\nabla f_{sdf}\|_2 = 1$. The weighted energy terms are

\begin{subequations} \label{eq:trajectory_terms}
\begin{align}
V_s(\mathbf{X}) &= \frac{w_s}{2}\sum^{N_t-1}_{t=2} \big\| \mathbf{x}_{t+1} \ominus \mathbf{x}_{t} - \mathbf{x}_{t} \ominus \mathbf{x}_{t-1} \big\|^{2}_{2}, \label{eq:Vs} \\
V_a(\mathbf{X}) &= \frac{w_a}{2} \sum^{N_t}_{t=1} \big\| \mathbf{x}_t(z)^{\top} \nabla f_{sdf} (\mathbf{x}_t) - 1 \big\|^2_2, \label{eq:Va} \\
V_f(\mathbf{X}) &= \frac{w_f}{2} \sum^{N_t}_{t=1} \big\| f_{sdf} (\mathbf{x}_t) \big\|_2^2, \label{eq:Vf} \\
V_e(\mathbf{X}) &= \frac{w_e}{2} \sum^{S}_{i=1} \Lambda_i (c(\mathbf{X})_i - \mathbf{c}_i)^2. \label{eq:Ve}
\end{align}
\end{subequations}
The smoothness term $V_s(\mathbf{X})$ is designed to minimize pose acceleration; The surface normal alignment energy $V_a(\mathbf{X})$ encourages the $z$-axis of the via-points to align with the normals of the SDF. In contrast, the surface attachment energy $V_f(\mathbf{X})$ ensures that the trajectory remains close to the SDF. Finally, the ergodic metric serves as the objective function introduced in Section~\ref{subsec:ergodic_prelim}. Note that all energy terms are constructed in a sum-of-squares form and can be written compactly as
$V(\mathbf{X})=\tfrac{1}{2}\, r(\mathbf{X})^{\top} r(\mathbf{X})$,
which enables Gauss--Newton (GN)-style preconditioning with
$\mathbf{H}=\mathbf{J}_r^{\top}\mathbf{J}_r$,
where $\mathbf{J}_r=\mathbf{J}r(\mathbf{X})$ can be derived in the same manner as in Eq.~(\ref{eq:transport_jacobian}). The structure of the Hessian for smoothness energy resembles that of the CHOMP planner \cite{zucker2013chomp}. When formulated on the Lie group, the induced smoothness metric remains coordinate-invariant, as in CHOMP; however, it is no longer constant. By directly applying the preconditioner with the gradient and retraction, one can obtain an efficient manifold-aware TO algorithm. We will evaluate it in section \ref{sec:experiment} (GN method). Therefore, the preconditioned SVGD component $\hat{\psi}$ can be implemented following the methods introduced in \cite{detommaso2018stein, wang2019stein}. 
For a set of $N_p$ trajectory particles with preconditioning matrices $\{\mathbf{H}_i \in \mathbb{R}^{6N_t \times 6 N_t}\}_{i=1}^{N_p}$, we adopt the block-diagonal preconditioning method proposed in \cite{detommaso2018stein}, the update direction is $\boldsymbol{\alpha} \in 
\mathbb{R}^{6 N_t}$ by solving the following linear system
\begin{equation}
    \left( \sum_{i=1}^{N_p} \mathbf{H}_i \, k(\mathbf{X}_i, \mathbf{X}')^2 + \mathrm{grad}_1\, k(\mathbf{X}_i, \mathbf{X}')^{\otimes 2} \right) \boldsymbol{\alpha} = \phi^*(\mathbf{X}')^{\vee},
\end{equation}
where the term $\mathrm{grad}_1\, k(\mathbf{X}_i, \mathbf{X}')^{\otimes 2}$ denotes the Euclidean outer product of the parallel transported kernel gradient $\mathcal{P}_{\mathbf{X}_i\rightarrow\mathbf{X}'}(\mathrm{grad}\, k(\mathbf{X}_i, \mathbf{X}')^{\vee}) \in \mathbb{R}^{6 N_t}$. This expression represents a direct extension of the approach proposed in \cite{detommaso2018stein}. The outer product of the kernel gradient can be interpreted as additional metric terms induced by the kernel structure. The kernel-weighted averaging of Gauss–Newton metrics provides a computational compromise necessary for algorithmic feasibility. Moreover, the computation can be further simplified through kernel design.

We adopt a sum of SE(3) kernels that compares the trajectory at each individual time step, which both simplifies the computation and mitigates potential performance degradation caused by high-dimensional particles.
\begin{equation}
    k(\mathbf{X}, \mathbf{X}') = \sum_{t=1}^{N_t} \exp\Bigg(-\frac{1}{l} \big\| \mathbf{x}_t \ominus \mathbf{x}'_t \big\|_2^2 \Bigg),
\end{equation}
where $l$ is the kernel length scale hyperparameter. All gradients can be efficiently computed by first performing automatic differentiation and then projecting onto the tangent space using the gradient in Eq.~(\ref{eq:riem_gradient}). The Jacobian of the residual can be computed analytically using the fundamental Jacobian of the Lie group and the chain rule; to avoid a cluttered explanation, we refer the reader to \cite{sola2018micro} for further technical details. To ensure strict constraint compliance, the current framework can be readily extended to an Augmented Lagrangian constrained optimization approach; however, for simplicity, we leave this for future work.
\section{EXPERIMENT}
\label{sec:experiment}
\subsection{Benchmark Test}

\begin{figure*}[t]
    \centering
    \includegraphics[width=\linewidth,height=0.5\linewidth]{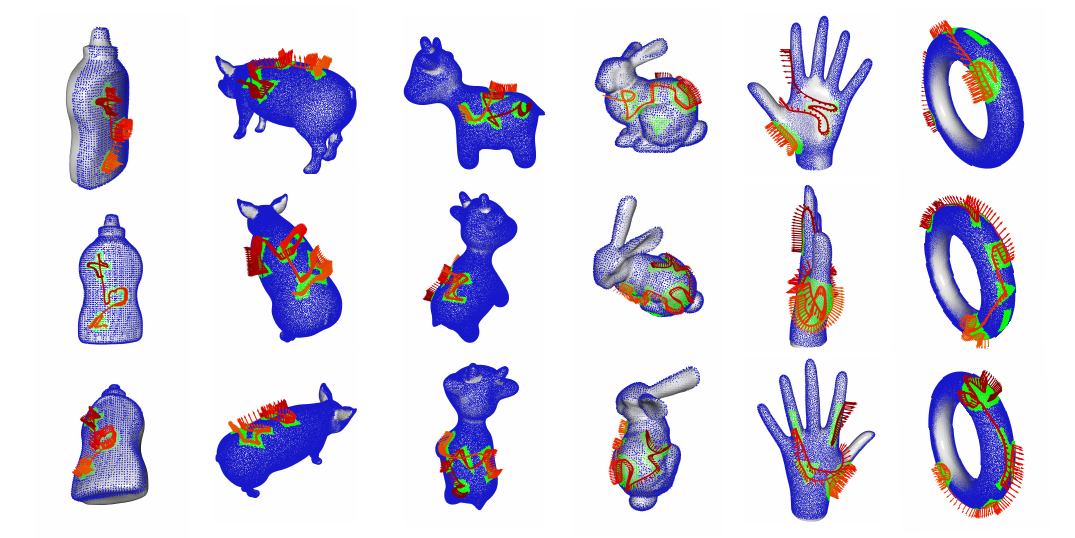}
    \caption{Six benchmark problems. From left to right: Mustard Bottle (simplified as Must), Pig, Spot, Bunny, Hand, and Torus. The background point cloud is shown in blue, ROIs in green, and the SDF in silver to enhance visualization. The TSVEC-planned trajectory is depicted as a solid line transitioning from light red to dark red, with normal vectors indicating time steps 1–200. Only the Z-axis is shown to avoid visual clutter. The objects are arranged according to ROI curvature variation, from flat to highly curved.}
    \vspace{-3mm}
    \label{fig:benchmark}
\end{figure*}

To evaluate the performance of our proposed ergodic TO algorithm, we designed a comprehensive benchmark using point cloud problems with colored point clouds as regions of interest (ROI), as shown in Fig.~\ref{fig:benchmark}. Each mesh was sourced from standard mesh datasets and uniformly sampled according to surface area to generate evenly distributed point clouds. The ROIs were intentionally positioned at distant locations to rigorously test the TO algorithms' effectiveness.

We conducted two groups of comparative experiments: single-point estimation methods (i.e., deterministic optimization that outputs a single trajectory/mode rather than a particle set) and particle-based methods. Within each group, all TO algorithms employed identical hyperparameters, initialization strategies, and shared random seeds to ensure fair comparison. To mitigate pathological optimization on a discrete point cloud that requires normal computation using K-nearest-neighbor, all TO methods utilized a neural SDF to represent object geometry. The neural SDF was implemented using an efficient JAX framework, enabling object training completion within 10 seconds. Ergodic metrics were computed directly on point clouds using a CUDA-accelerated batched nearest neighbor package (JAXkd). All experiments were conducted on a single RTX 4080 GPU using single-precision arithmetic for compatibility with the CUDA-accelerated point cloud KNN implementation (K=60).

For the single point estimation group: We selected LBFGS from the Optax library with backtracking line search as the off-the-shelf optimizer for unconstrained TO. Gradients were computed using JAX autodiff, with quaternion components projected to the unit sphere after each update, serving as a vanilla baseline unaware of the manifold structure. We also employed the IPOPT optimizer via its official Python wrapper, coupled with the autodiff pipeline to represent second-order TO in linearized coordinates. SE(3) trajectories were parameterized in Lie algebra space and mapped back to the manifold via the exponential map for objective function evaluation. The CHOMP-like TO algorithm proposed in Section \ref{subsec:se_3_to} was implemented in JAX and denoted as GN, which uses backtracking line search and adaptive damping as globalization.

For the particle-based group: We implemented our method, TSVEC, as well as the vanilla Stein ergodic TO introduced in \cite{lee2024stein}, denoted as Stein Ergodic (SE), using JAX. To make SE work on SE(3), we adopted the SE(3) gradient, retraction, and parallel transport mechanisms. To showcase that the performance gain is not due to accidentally better initialization, we implemented Batch GN, which runs a batch of parallel GN as in single-point estimation without the Stein variational mechanism.

The trajectory length $N_t$ is 200 steps, generating an optimization problem with 1400 decision variables. The weights assigned in $V_s, V_a, V_f,$ and $V_e$ are $5.0, 3.0, 3.0,$ and $0.1$, respectively. To distribute the effect of the trajectory realistically, we apply a Gaussian smoothing kernel, simulating the agents' radius. The stopping criterion was $1e^{-4}$ for single precision compatibility. We used straight line initialization for all methods connecting the furthest two colored point clouds, as this might be the only prior knowledge available about the scenario. The start poses were initialized and aligned with the world coordinate frame. For particle-based methods, we added small white noise with 0.005 variance to each via point, as SVGD-based optimizers generate infinitely large kernel gradients if two particles overlap. We use point cloud diffusion to generate a globally smooth vector field, convexly combined with local sharp features to guide the optimization. The number of principal components was set to 300 to capture complex surface patterns.

The theoretically sound merit function for SVGD line search or stopping criteria is the kernelized Stein discrepancy (KSD) discussed in \cite{liu2016kernelized}, which provides a principled way to quantify how far the particle distribution deviates from the target in kernel space. However, KSD evaluation is expensive and numerically unstable due to its dependence on higher-order derivatives. Therefore, no reasonable stopping criterion exists for TSVEC and SE. Thus, we ran them for maximum iterations: 1000 for TSVEC and 9999 for SE. Since vanilla SE suffers from convergence issues, we allowed it ten times more iterations. The step size for TSVEC is 0.1, and 0.02 for SE. All particle methods used 100 particles. We tested this within our framework and found that using 500 particles does not incur memory issues on a single GPU. Kernel lengths were set to 0.05, except for the torus (0.1), as it needs more exploration. We found that the mean adaptive kernel-length heuristic was less effective than expected, and therefore removed it from the current framework. Results for methods tested on the benchmark were run 10 times, and we reported their mean values in TABLE \ref{tab:method_comparison}. We do not emphasize computation time as a criterion, since IPOPT requires dense Hessian computation via automatic differentiation and is therefore extremely slow. Instead, we use the iteration count and the final objective value as the key evaluation criteria. A dash (‘–’) indicates method failure.

The table is organized by the curvature variation of the scenario surface. As shown, L-BFGS rarely achieves satisfactory performance, with direct quaternion clipping causing premature termination. While IPOPT performs considerably better, it outperforms the GN planner only once in the low surface curvature case. GN consistently surpasses both baselines in convergence iterations and final objective values. Moreover, Batch GN consistently outperforms single GN, which aligns with expectations. Most notably, TSVEC consistently achieves superior performance across all scenarios, obtaining significantly better objective values than all competing approaches. In contrast, SE suffers from convergence difficulties, as the 200-step trajectory results in a severely ill-conditioned optimization problem. During experimentation, we observed that while the ergodic objective's mathematical form remains unchanged, it exhibits slight non-smoothness due to the nearest-neighbor search required for its evaluation.

\begin{table}[htpb]
\centering
\caption{Performance comparison across benchmark scenarios.}
\label{tab:method_comparison}
\setlength{\tabcolsep}{1.pt}
\renewcommand{\arraystretch}{0.7}
\begin{tabular}{llccccccc}
\toprule
Scen. & Method & $V_s$ & $V_a$ & $V_f$ & $V_e$ & \cellcolor{gray!20}$V$ & \cellcolor{gray!20}Iter. & Time(s) \\
\midrule
\multirow{6}{*}{Must} 
& LBFGS & 2.26e-2 & 7.02e-4 & 2.31e-4 & 1.87e-1 & \cellcolor{gray!20}2.10e-1 & \cellcolor{gray!20}153 & 3.16 \\
& IPOPT & 1.67e-2 & 9.48e-3 & 1.06e-3 & 1.07e-1 & \cellcolor{gray!20}\textbf{1.34e-1} & \cellcolor{gray!20}1000 & 600 \\
& GN  & 7.84e-3 & 1.67e-3 & 9.82e-4 & 1.41e-1 & \cellcolor{gray!20}1.51e-1 & \cellcolor{gray!20}\textbf{58} & \textbf{0.392} \\
\cmidrule{2-9}
& SE & 1.37e-1 & 2.23e+0 & 2.98e-2 & 3.86e-1 & \cellcolor{gray!20}2.79e+0 & \cellcolor{gray!20}9999 & 220 \\
& Batch GN  & 6.32e-3 & 1.81e-3 & 8.84e-4 & 1.24e-1 & \cellcolor{gray!20}1.33e-1 & \cellcolor{gray!20}116 & 12.1 \\
& TSVEC  & 1.78e-3 & 1.98e-4 & 2.12e-4 & 8.88e-2 & \cellcolor{gray!20}\textbf{9.10e-2} & \cellcolor{gray!20}1000 & 138 \\
\midrule
\multirow{6}{*}{Pig}
& LBFGS & -& -& -& -& \cellcolor{gray!20}- & \cellcolor{gray!20}- & - \\
& IPOPT & 3.37e-3 & 4.87e-3 & 1.62e-4 & 3.95e-1 & \cellcolor{gray!20}4.03e-1 & \cellcolor{gray!20}499 & 128 \\
 & GN  & 3.29e-3 & 6.28e-4 & 4.57e-5 & 2.97e-2 & \cellcolor{gray!20}\textbf{3.37e-2} & \cellcolor{gray!20}\textbf{52} & \textbf{0.441} \\
\cmidrule{2-9}
& SE & 1.47e-2 & 1.63e-1 & 6.35e-3 & 6.18e-1 & \cellcolor{gray!20}8.02e-1 & \cellcolor{gray!20}9999 & 249 \\
& Batch GN  & 2.69e-3 & 3.97e-4 & 4.51e-5 & 2.68e-2 & \cellcolor{gray!20}2.99e-2 & \cellcolor{gray!20}123 & 14.0 \\
& TSVEC  & 1.16e-3 & 1.33e-4 & 7.68e-6 & 9.42e-3 & \cellcolor{gray!20}\textbf{1.07e-2} & \cellcolor{gray!20}1000 & 145 \\
\midrule
\multirow{6}{*}{Spot} 
& LBFGS & 1.97e-2 & 3.32e-2 & 5.44e-3 & 5.11e-1 & \cellcolor{gray!20}5.69e-1 & \cellcolor{gray!20}137 & 9.58 \\
& IPOPT & 5.20e-3 & 6.94e-3 & 4.28e-4 & 3.16e-1 & \cellcolor{gray!20}3.28e-1 & \cellcolor{gray!20}1000 & 290 \\
& GN & 2.04e-3 & 7.91e-4 & 4.02e-5 & 1.87e-1 & \cellcolor{gray!20}\textbf{1.90e-1} & \cellcolor{gray!20}\textbf{40} & \textbf{0.359} \\
\cmidrule{2-9}
& SE & 1.22e-1 & 8.45e-1 & 9.57e-3 & 1.28e+0 & \cellcolor{gray!20}2.26e+0 & \cellcolor{gray!20}9999 & 244 \\
& Batch GN  & 1.58e-3 & 3.84e-4 & 2.29e-5 & 1.79e-2 & \cellcolor{gray!20}1.99e-2 & \cellcolor{gray!20}119 & 13.6 \\
& TSVEC & 1.17e-3 & 2.54e-4 & 9.04e-6 & 6.86e-3 & \cellcolor{gray!20}\textbf{8.29e-3} & \cellcolor{gray!20}1000 & 144 \\
\midrule
\multirow{6}{*}{Bunny} 
& LBFGS & 7.21e-3 & 6.24e-3 & 7.15e-4 & 1.34e-1 & \cellcolor{gray!20}1.48e-1 & \cellcolor{gray!20}262 & 3.88 \\
& IPOPT & 5.42e-3 & 4.50e-3 & 5.15e-4 & 1.16e-1 & \cellcolor{gray!20}1.26e-1 & \cellcolor{gray!20}199 & 38.8 \\
& GN  & 5.56e-3 & 2.69e-3 & 3.73e-4 & 7.74e-2 & \cellcolor{gray!20}\textbf{8.60e-2} & \cellcolor{gray!20}\textbf{36} & \textbf{0.317} \\
\cmidrule{2-9}
& SE & 2.06e-2 & 8.98e-2 & 4.44e-3 & 2.44e-1 & \cellcolor{gray!20}3.59e-1 & \cellcolor{gray!20}9999 & 224 \\
& Batch GN  & 4.44e-3 & 2.00e-3 & 2.61e-4 & 3.65e-2 & \cellcolor{gray!20}4.32e-2 & \cellcolor{gray!20}107 & 11.5 \\
& TSVEC  & 3.81e-3 & 1.59e-3 & 2.19e-4 & 1.78e-2 & \cellcolor{gray!20}\textbf{2.34e-2} & \cellcolor{gray!20}1000 & 138 \\
\midrule
\multirow{6}{*}{Hand} 
& LBFGS & 7.34e-3 & 5.10e-3 & 1.38e-3 & 5.22e-1 & \cellcolor{gray!20}5.36e-1 & \cellcolor{gray!20}70 & 3.04 \\
& IPOPT & 3.61e-3 & 2.16e-3 & 2.88e-4 & 3.07e-1 & \cellcolor{gray!20}3.13e-1 & \cellcolor{gray!20}256 & 61.3 \\
& GN  & 1.01e-2 & 2.29e-3 & 3.12e-3 & 1.50e-1 & \cellcolor{gray!20}\textbf{1.66e-1} & \cellcolor{gray!20}\textbf{49} & \textbf{0.412} \\
\cmidrule{2-9}
& SE & 2.39e-2 & 1.86e-2 & 2.04e-3 & 6.79e-1 & \cellcolor{gray!20}7.24e-1 & \cellcolor{gray!20}9999 & 230 \\
& Batch GN  & 6.76e-3 & 1.51e-3 & 5.08e-4 & 9.97e-2 & \cellcolor{gray!20}1.08e-1 & \cellcolor{gray!20}150 & 16.3 \\
& TSVEC  & 4.49e-3 & 6.81e-4 & 2.83e-4 & 6.83e-2 & \cellcolor{gray!20}\textbf{7.38e-2} & \cellcolor{gray!20}1000 & 139 \\
\midrule
\multirow{6}{*}{Torus} 
& LBFGS & 3.71e-2 & 1.95e-2 & 1.98e-3 & 7.41e-1 & \cellcolor{gray!20}8.00e-1 & \cellcolor{gray!20}114 & 9.47 \\
& IPOPT & 9.31e-3 & 1.26e-2 & 1.03e-3 & 6.39e-1 & \cellcolor{gray!20}6.62e-1 & \cellcolor{gray!20}1000 & 274 \\
& GN  & 3.66e-3 & 5.56e-4 & 9.50e-5 & 1.78e-1 & \cellcolor{gray!20}\textbf{1.83e-1} & \cellcolor{gray!20}\textbf{60} & \textbf{0.525} \\
\cmidrule{2-9}
& SE & 5.87e-2 & 4.44e-1 & 1.72e-2 & 9.16e-1 & \cellcolor{gray!20}1.44e+0 & \cellcolor{gray!20}9999 & 254 \\
& Batch GN  & 3.09e-3 & 4.58e-4 & 5.51e-5 & 8.74e-2 & \cellcolor{gray!20}9.10e-2 & \cellcolor{gray!20}294 & 32.6 \\
& TSVEC  & 2.99e-3 & 6.88e-4 & 2.63e-4 & 5.68e-2 & \cellcolor{gray!20}\textbf{6.08e-2} & \cellcolor{gray!20}1000 & 140 \\
\bottomrule
\end{tabular}
\end{table}

\subsection{Real World Task}
To evaluate TSVEC's performance in real-world surface manipulation, we deployed a Franka Emika Panda robot to track trajectories corresponding to SE(3) motions planned by our framework. The target point cloud represents hand-written random geometries and characters, obtained by sampling from the mesh and colored using a rendering tool. A cooking pot with a 22 cm diameter and 16 cm height was used to simulate a cylindrical curved surface. To execute the task, we first converted TSVEC task-space trajectories into joint-space commands using a sequential MuJoCo inverse-kinematics (IK) planner with damped least squares, warm-started from the previous time step’s solution. The resulting joint trajectories were then executed on the robot via joint-space position control. The real-world tracking results are presented in Fig. \ref{fig:real_world_panda}.

\begin{figure}[tpb]
    \centering
    \includegraphics[width=0.99\linewidth,height=0.4\linewidth]{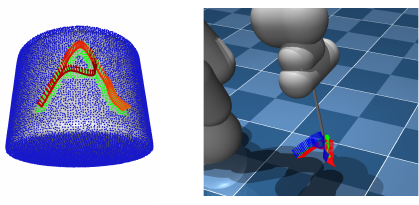} 
    \caption{Left: SE(3) planning result on the hand-colored point cloud surface. Right: the corresponding sequential IK solution. The viewing angle matches that of Fig.~\ref{fig:real_world_panda}, allowing a direct comparison between the planned and real-world results.}
    \vspace{-2mm}
    \label{fig:real_world}
\end{figure}

When applied to the drawing task, the GN planner is largely unable to produce recognizable characters, as it frequently becomes trapped in local minima and lacks sufficient exploration. This limitation is particularly critical in ergodic trajectory optimization, where broader exploration is necessary to discover better suboptimal solutions; instead, GN trajectories tend to converge directly to poor local minima. In contrast, TSVEC generates trajectories that are mostly recognizable and substantially reduces the number of trials required. An example of the planning result for the letter "A" is shown in Fig.~\ref{fig:real_world}.
\section{CONCLUSIONS}
This work introduced TSVEC, a preconditioned SE(3) SVGD framework for ergodic trajectory optimization on discrete point cloud surfaces. By casting the problem as manifold-aware sampling-as-optimization, our method addresses a key difficulty of local gradient-based ergodic trajectory optimization on point cloud targets, namely its tendency to become trapped in poor local minima under highly non-convex surface distributions.

The proposed approach combines SE(3)-aware Stein variational updates, which maintain geometric consistency in particle interactions, with Gauss–Newton-style preconditioning that improves the conditioning of long-horizon optimization problems. Together, these design choices make the resulting SAO framework substantially more robust to the trivial local optima that often affect optimization-based alternatives. Across the benchmark scenarios considered in this paper, TSVEC consistently achieved better coverage quality and lower final objective values than strong baselines, including IPOPT, L-BFGS, and SE(3)-aware Gauss–Newton methods.

In the real-world surface drawing experiments, TSVEC further produced recognizable character trajectories with little hand tuning, whereas the optimization-based planners we tested often converged prematurely to trajectories with poor visual structure. Taken together, these results show that TSVEC is an effective and robust method for ergodic trajectory optimization on discrete point cloud surfaces, particularly in settings where non-convexity and SE(3) geometric constraints make purely local optimization difficult.

\addtolength{\textheight}{-0.5cm}   







\bibliographystyle{IEEEtran}
\bibliography{reference} 


\end{document}